\documentclass[letterpaper, 10pt, conference]{ieeeconf}      

\IEEEoverridecommandlockouts                              

\usepackage{subcaption}
\usepackage[utf8]{inputenc}
\usepackage[english]{babel}

\usepackage{amsmath}
\usepackage{multicol}
\usepackage{mathtools}
\usepackage[table]{xcolor}
\usepackage[verbose]{wrapfig}
\usepackage{url}
\usepackage{svg}
\usepackage{caption}
\usepackage{threeparttable}
\usepackage{graphicx}
\usepackage{array}
\usepackage{hhline}
\usepackage{breqn}
\usepackage{soul}
\usepackage{amssymb}
\usepackage{cite}
\DeclareMathOperator*{\argmax}{arg\,max}

\newcolumntype{P}[1]{>{\centering\arraybackslash}p{#1}}
 
\title{\LARGE \bf
Perimeter Control Using Deep Reinforcement Learning: A Model-free Approach towards Homogeneous Flow Rate Optimization
}

\author{Xiaocan Li$^{1}$, Ray Coden Mercurius$^{2}$, Ayal Taitler$^{1}$, Xiaoyu Wang$^{2}$, Mohammad Noaeen$^{3}$, \\ Scott Sanner$^{1}$, and Baher Abdulhai$^{2}$
\thanks{$^{1}$Xiaocan Li, Ayal Taitler and Scott Sanner are with the Department of Mechanical \& Industrial Engineering, University of Toronto, Canada
        {\tt\small hsiaotsan.li@mail.utoronto.ca, ataitler@gmail.com, ssanner@mie.utoronto.ca}}%
\thanks{$^{2}$Ray Coden Mercurius, Xiaoyu Wang, Baher Abdulhai are with the Department of Civil \& Mineral Engineering, University of Toronto, Canada
        {\tt\small ray.mercurius@mail.utoronto.ca, cnxiaoyu.wang@mail.utoronto.ca, baher.abdulhai@utoronto.ca}}%
\thanks{$^{3}$Mohammad Noaeen is with the Dalla Lana School of Public Health, University of Toronto, Canada
        {\tt\small m.noaeen@utoronto.ca}}%
}

\begin{document}

\maketitle
\thispagestyle{empty}
\pagestyle{empty}

\begin{abstract}
Perimeter control maintains high traffic efficiency within protected regions by controlling transfer flows among regions to ensure that their traffic densities are below critical values. Existing approaches can be categorized as either model-based or model-free, depending on whether they rely on network transmission models (NTMs) and macroscopic fundamental diagrams (MFDs). Although model-based approaches are more data efficient and have performance guarantees, they are inherently prone to model bias and inaccuracy. For example, NTMs often become imprecise for a large number of protected regions, and MFDs can exhibit scatter and hysteresis that are not captured in existing model-based works. Moreover, no existing studies have employed reinforcement learning for homogeneous flow rate optimization in microscopic simulation, where spatial characteristics, vehicle-level information, and metering realizations --- often overlooked in macroscopic simulations --- are taken into account. To circumvent issues of model-based approaches and macroscopic simulation, we propose a model-free deep reinforcement learning approach that optimizes the flow rate homogeneously at the perimeter at the microscopic level. Results demonstrate that our model-free reinforcement learning approach without any knowledge of NTMs or MFDs can compete and match the performance of a model-based approach, and exhibits enhanced generalizability and scalability.
\end{abstract}

\section{Introduction}

Perimeter control manages inter-regional transfer flows by gating the feeder links of protected regions, ensuring that each region's vehicle accumulation does not surpass a critical value that would negatively impact traffic efficiency. Urban traffic networks frequently become oversaturated during rush hour, necessitating perimeter control. The delineation of a congested area is called perimeter identification, which can be executed statically or dynamically \cite{guo2019dynamic}. Once the perimeter is identified, perimeter control operates at the regional level \cite{ren2020data, li2021perimeter} and is particularly effective in oversaturated situations where adaptive traffic signal control alone is insufficient \cite{keyvan2019traffic}.



Recent works can be divided into reinforcement learning (RL) based control and non-RL-based control. Non-RL control comprises classical feedback control methods, such as Proportional Integral (PI) controller \cite{keyvan2012exploiting, keyvan2015controller}, which aims to maintain densities near set-points determined by MFDs. Analytical control approaches such as optimal control \cite{aalipour2018analytical}, model predictive control \cite{ramezani2015dynamics}, and switching algorithm on top of a linear quadratic regulator \cite{li2021perimeter} require an NTM and/or an MFD. Adaptive and learning control methods such as \cite{kouvelas2017enhancing, ren2020data, su-neuro-dynamic2020} need a model for tuning and are limited by the policy class they can approximate.
For RL-based control, \cite{NI2019358} employs reinforcement learning to heterogeneously redistribute a predetermined total regional flow rate among different feeder links at the microscopic level. However, the prior knowledge of the total rate is still obtained from a model-based non-RL controller. In \cite{zhou2021modelfree}, a deep RL approach is utilized to regulate the transfer flow, using the NTM as a plant to facilitate the macroscopic environment. However, the macroscopic simulation has its limitations. 

The macroscopic simulation based on the NTM lacks consideration for: 1) spatial characteristics, which include road geometry, the spatial distribution of congestion, and traffic network topology; 2) vehicle-level information, which comprises stochastic route choices, driving aggressiveness, physical attributes such as car length and width, and generation times; 3) most importantly, metering realizations, such as traffic lights and flow metering, cannot be simulated at a macroscopic level, even though they are crucial for field implementation. Macroscopic simulations do not account for these factors, resulting in an environment that deviates from reality. Consequently, we are motivated to propose a model-free controller that learns through interaction with a microscopic environment that are more detailed and realistic.

In this work, we formulate perimeter control as a model-free RL problem, employing the Proximal Policy Optimization (PPO) algorithm. To validate the proposed method, we create a traffic network and demand profile in a microscopic traffic simulator, allowing the RL agent to interact and collect training and evaluation data. Then, we provide analysis of state designs,  policy visualization, and various traffic metrics. Finally, we conduct a comprehensive evaluation of different demand scenarios for the PPO and baseline controllers.

Our contributions are the following: 1) to the best of our knowledge, we are the first to propose a model-free deep RL approach for homogeneous flow rate optimization of perimeter control at the microscopic level; 2) the incorporation of microscopic environment facilitates a trustworthy simulation for future field implementation; 3) extensive analysis of the proposed approach's robustness and generalizability towards unseen demand scenarios is provided.

\section{Problem Statement}
\begin{figure}[!htbp]
    \centering
    \includegraphics[width=0.3\textwidth]{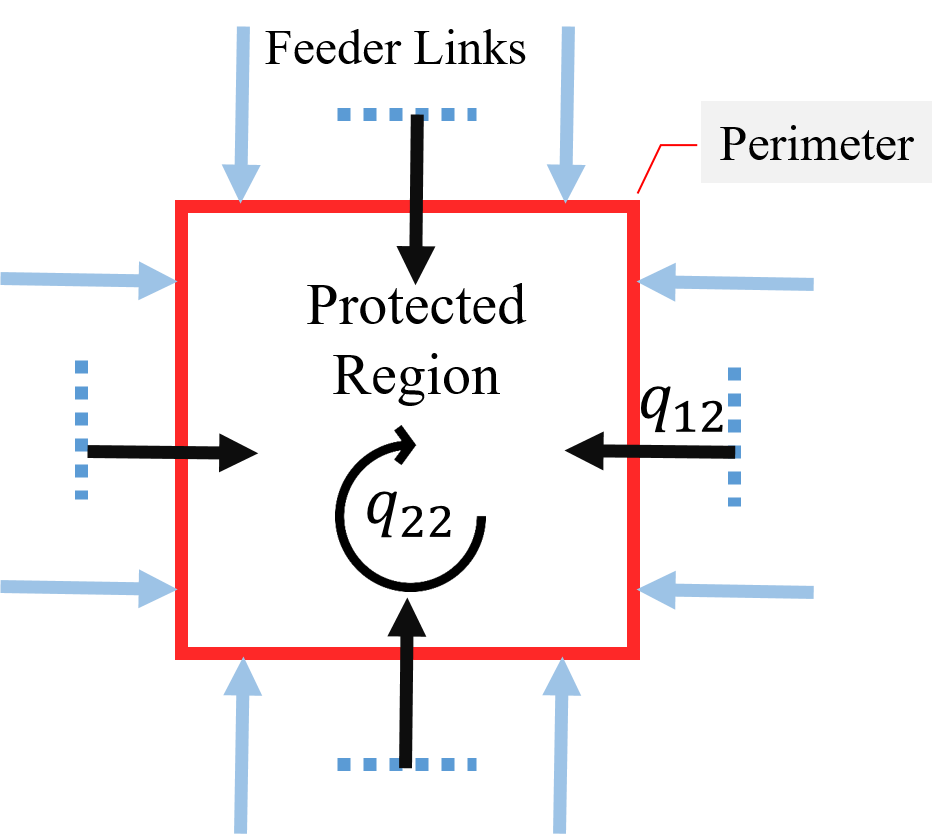}
    \caption{A generic traffic network with single protected region.}
    \label{fig:generic-network}
\end{figure}

A single protected region perimeter control problem is formulated for its simplicity. In Figure \ref{fig:generic-network}, the abstract traffic network consists of one protected region and several feeder links. A fixed control plan of traffic signals is used for intersections in the protected region. During high traffic demand periods, such as morning rush hour, incoming vehicles from feeder links are regulated by meterings to protect the region. Only incoming vehicles are restricted, while outgoing vehicles are unaffected. At each control timestep, the controller sets the allowed flow rate for each flow metering. To simplify the problem, we assume a homogeneous perimeter control, implying that all feeder links employ an identical flow metering rate. 
Flow meterings are used instead of intersection timings for gating due to their simpler control nature, which eases the complexity of validation.

A generic demand profile with a high peak of trips exceeding the protected region's capacity is required. Destinations are within the protected region, while origins are in both the protected region and feeder links. Demand flows are denoted by $q_{ij} (veh/s), i,j\in\{1,2\}$, with 1 representing  links outside the protected region and 2 representing the protected region. Therefore, $q_{12}$ denotes flow from feeder links to the protected region, also known as exogenous flow; $q_{22}$ denotes endogenous flow within the protected region; $q_{21}$ denotes flow from protected region to exit links; and $q_{11}$ denotes flow from feeder links to exit links. To simplify the demand profile, $q_{11}$ and $q_{21}$ are set to zero, as is in \cite{NI2019358}.

The objective of perimeter control is to minimize total delay. However, measuring delay is challenging due to difficulties in capturing stopping time and speed decrease for each vehicle. Instead, total time spent (TTS) is used as a surrogate, calculated as the difference between trip exit and generation times. Since TTS comprises both delay and travel time without delay, minimizing TTS is equivalent to minimizing total delay.


\section{Methodology}
\subsection{Formulation as Markov Decision Processes}
The perimeter control problem is formulated as a model-free reinforcement learning problem: At each timestep $t$, an agent takes an action $a_t$ based on the current state $s_t$ of an environment and receives a scalar reward $r_{t+1}$ from the environment. The reinforcement learning is formalized by the Markov Decision Process (MDP)  described by a tuple $(\mathcal{S}, \mathcal{A}, \mathcal{T}, \mathcal{R}, \gamma)$:

1) State space $\mathcal{S}$: The design of the state is crucial for developing an agent that is robust and generalizable across various scenarios. Three traffic metrics serve as state variables: \textit{I)} inner density $D_p$ represents protected region congestion level defined as Eq. (\ref{eq:inner-density}); \textit{II)} feeder link density $D_f$  defined as Eq. (\ref{eq:feeder-link-density}) is correlated with queue length outside the perimeter; \textit{III)} future demand $n_{ij}, i,j\in \{1,2
\}$, derived from historical data, denotes the number of vehicles generated in the next control cycle $T$, defined as Eq. (\ref{eq:future-demand}). Variables $L_i, V_i$ represent the length of link $i$ and the number of vehicles on link $i$, and $\mathcal{P}, \mathcal{F}$ are the set of links for the protected region and feeder links, respectively. Three candidate designs of state space to be validated are defined in Eq. (\ref{eq:state1}), (\ref{eq:state2}), (\ref{eq:state3}), where each design contains more information than previous ones. The definition in Eq. (\ref{eq:state3}) is adopted from \cite{zhou2021modelfree}, while Eq. (\ref{eq:state1}), (\ref{eq:state2}) are our novel designs.
Utilizing future demand suggests that the controller has access to a demand prediction model, albeit at a significantly lower level than a comprehensive model of environment dynamics;


\begin{align}
D_p &= \sum_{i\in\mathcal{P}} V_i / \sum_{i\in\mathcal{P}} L_i \label{eq:inner-density} \\
D_f &= \sum_{i\in\mathcal{F}} V_i / \sum_{i\in\mathcal{F}} L_i\label{eq:feeder-link-density}\\
n_{ij}(t) &= \int_{t}^{t+T} q_{ij}(t') \, dt', \qquad i,j\in\{1,2\}\label{eq:future-demand} \\
s_t^{1} &= \begin{bmatrix} D_p \end{bmatrix} \label{eq:state1} \\
s_t^{2} &= \begin{bmatrix} D_p & D_f \end{bmatrix}^\top\label{eq:state2} \\
s_t^{3} &= \begin{bmatrix} D_p & D_f & n_{12} & n_{22} & n_{21} & n_{11} \end{bmatrix}^\top \label{eq:state3}
\end{align}

2) Action space $\mathcal{A}$: The action is defined as the flow metering rate, which represents the number of vehicles allowed to pass the meter within a unit of time. The action space is one-dimensional due to the homogeneous control assumption. For practicality, the applied action is bounded by a minimum of 50 $veh/h$ and a maximum of 300 $veh/h$ per meter;

3) Transition dynamics: The dynamics from state $s_t$ to new state $s_{t+1}$ by taking action $a_t$. In a model-free setting, the dynamics is provided by the traffic simulator Aimsun and remains unknown to the agent;

4) Reward $\mathcal{R}$: The trip completion rate ($veh/s$) for each control cycle as a measure of traffic efficiency;

5) Discount factor $\gamma$: A scalar value ranged in [0, 1] used to prioritize the current reward over future rewards. For our specific reward, $\gamma=1$ cannot further stimulate the agent to learn when all trips are completed, as the total 'discounted' reward equals a constant of the total number of trips completed. In contrast, $\gamma < 1$ encourages the agent to complete all trips earlier;



The state value $V_{\pi}(s)$ defined in Eq. (\ref{eq:v-value}) as the expected cumulative discounted reward over horizon $H$ under policy $\pi$. The horizon is finite as the simulated demand shown in Figure \ref{fig:demand-profile} is time-limited. The objective of RL is to maximize the expected state value over a particular state distribution $\mu$, and the optimal policy is found in Eq. (\ref{eq:optimal-policy}).
\begin{align}
V_\pi(s) &= \mathbb{E}_\pi[\sum_{t=0}^{H-1} \gamma^t r_t \mid s_t=s]\label{eq:v-value} \\
\pi^* &= \argmax_{\pi} \mathbb{E}_{s\sim\mu}\left[V_\pi(s)\right] \label{eq:optimal-policy}
\end{align}

\subsection{Proximal Policy Optimization}
Proximal Policy Optimization \cite{schulman2017proximal} is chosen as our RL algorithm due to its well-established reputation and ability to handle continuous action spaces such as flow metering rate. PPO is a policy gradient method that consists of an actor and a critic network. The actor network receives the state and generates action distributions, while the critic network receives the state and evaluates the state value. With these features, PPO is suitable for learning the optimal control policy in a perimeter control setting.

\section{Empirical Results \& Discussion}
\subsection{Experimental Setup}

\begin{figure}[!htbp]
    \centering
    \includegraphics[width=0.3\textwidth]{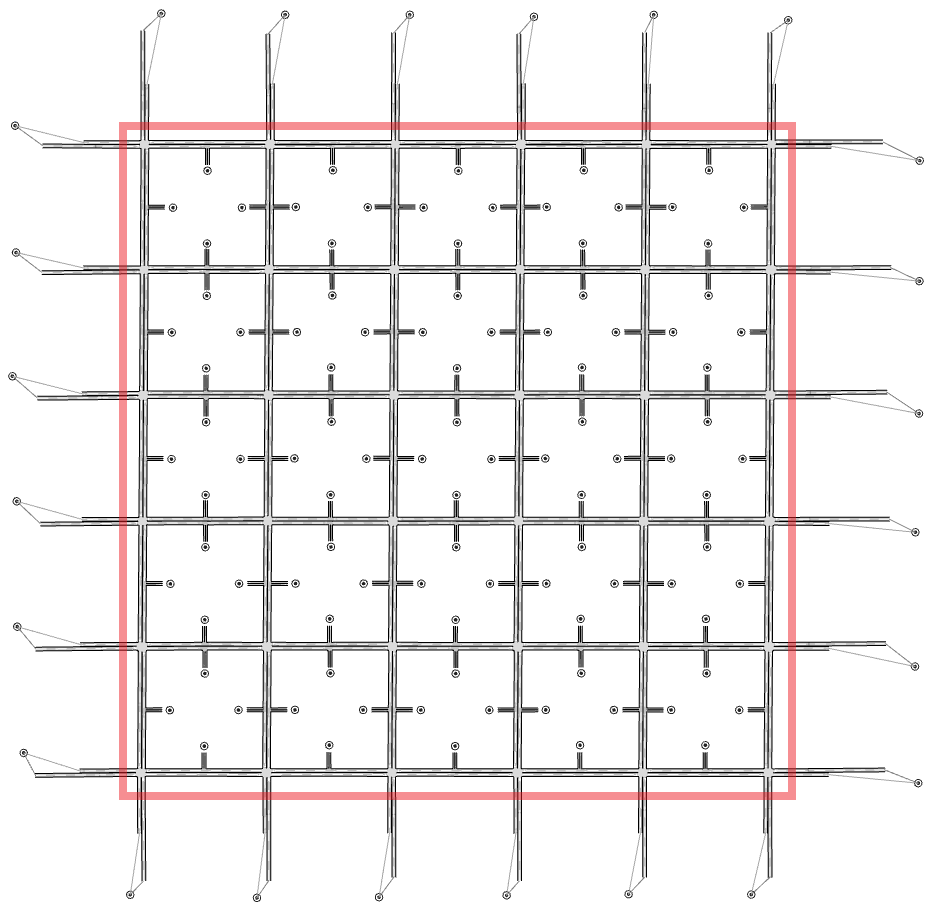}
    \caption{Testbed: $5\times5$ grid network with single protected region and 24 feeder links.}
    \label{fig:experiment-network}
\end{figure}

\textbf{Traffic Simulation Setup}: Figure \ref{fig:experiment-network} depicts the testbed traffic network consisting of a protected region with 24 feeder links. Each traffic link is 170 meters long with two lanes per link, except for links connected to an origin or destination within the protected region  have only one lane. In terms of the spatial distribution of origins and destinations, 100 points inside the protected region serve as both origins and destinations, while 24 points connected to each feeder link serve as origins only as $q_{21}=0$ and $q_{11}=0$.

\begin{figure}[!htbp]
    \centering
    \includegraphics[width=0.4\textwidth]{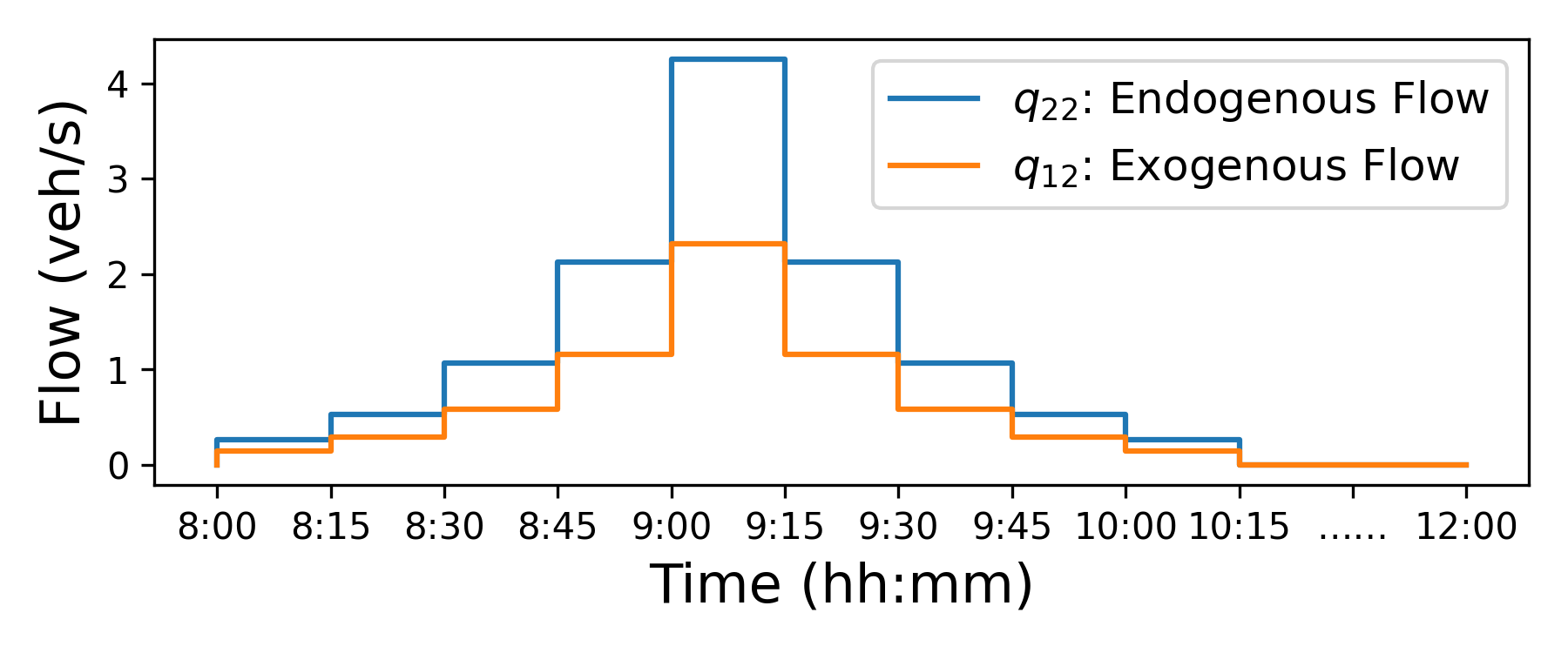}
    \caption{Traffic demand profile.}
    \label{fig:demand-profile}
\end{figure}

In Figure \ref{fig:demand-profile}, the traffic demand spanning 135 minutes is divided into nine 15-minute intervals. All origin-destination pairs share the same number of trips at each timestep, while the numbers of trips for nine intervals are determined by a vector of weights: $[r^0, r^1, r^2, r^3, r^4, r^3, r^2, r^1, r^0]$, where $r=2$ serves as a hyperparameter to regulate the peakedness of the demand profile. The same weights are applied to both a total endogenous demand of 11000 vehicles and a total exogenous demand of 6000 vehicles.


During training, the random seed is fixed to 1 for a stable environment. However, during evaluation, distinct random seeds are used to assess the controller's performance more comprehensively and robustly under varying conditions.

\textbf{RL Setup}: The control cycle $T$ is 96 seconds, same as the cycle of traffic signals in the protected region. Table \ref{tab:ppo_parameters} provides a comprehensive summary of PPO algorithm parameters and architecture for both actor and critic networks and key hyperparameters. State variables and rewards are scaled to the range $[0,1]$ to ensure numerical stability. 


\begin{table}[h!]
\centering
\caption{Summary of PPO Parameters.}
\begin{tabular}{|l|l|}
\hline
\textbf{Parameter} & \textbf{Value} \\ \hline
Actor Input Layer & $|S|=1,2,3$ units \\ \hline
Actor Hidden Layers & [256, 256] (Tanh activation) \\ \hline
Actor Output Layer & $|A|=1$ (ReLU activation) \\ \hline
Critic Input Layer & $|S|=1,2,3$ units \\ \hline
Critic Hidden Layers & [256, 256] (Tanh activation) \\ \hline
Critic Output Layer & 1 (No activation) \\ \hline
Training Episodes & 1200 \\ \hline
Batch Size & 25 episodes \\ \hline
Learning Rate & 5$e$-4 \\ \hline
Entropy Coefficient & 0.01 \\ \hline
Discount Factor $\gamma$ & 0.95, 0.9, 0.85 \\ \hline
\end{tabular}
\label{tab:ppo_parameters}
\end{table}


\subsection{Simulation Results}
Three perimeter controllers are evaluated: No Perimeter Control (NPC) as a naïve baseline, PI controller as a model-based baseline, and our proposed PPO controller. Comparative analysis between NPC and PI controllers highlights the importance of perimeter control, while comparative analysis between PI and PPO emphasizes the benefits of the model-free RL approach. Evaluations are conducted using 10 replications with random seeds from 1 to 10, accounting for the inherent stochasticity of microscopic simulations. The total number of vehicles remains consistent across replications, with a maximum difference of 4\%.



\begin{figure}[htbp]
\centering
  \begin{subfigure}[b]{0.48\columnwidth}
    \includegraphics[width=\linewidth]{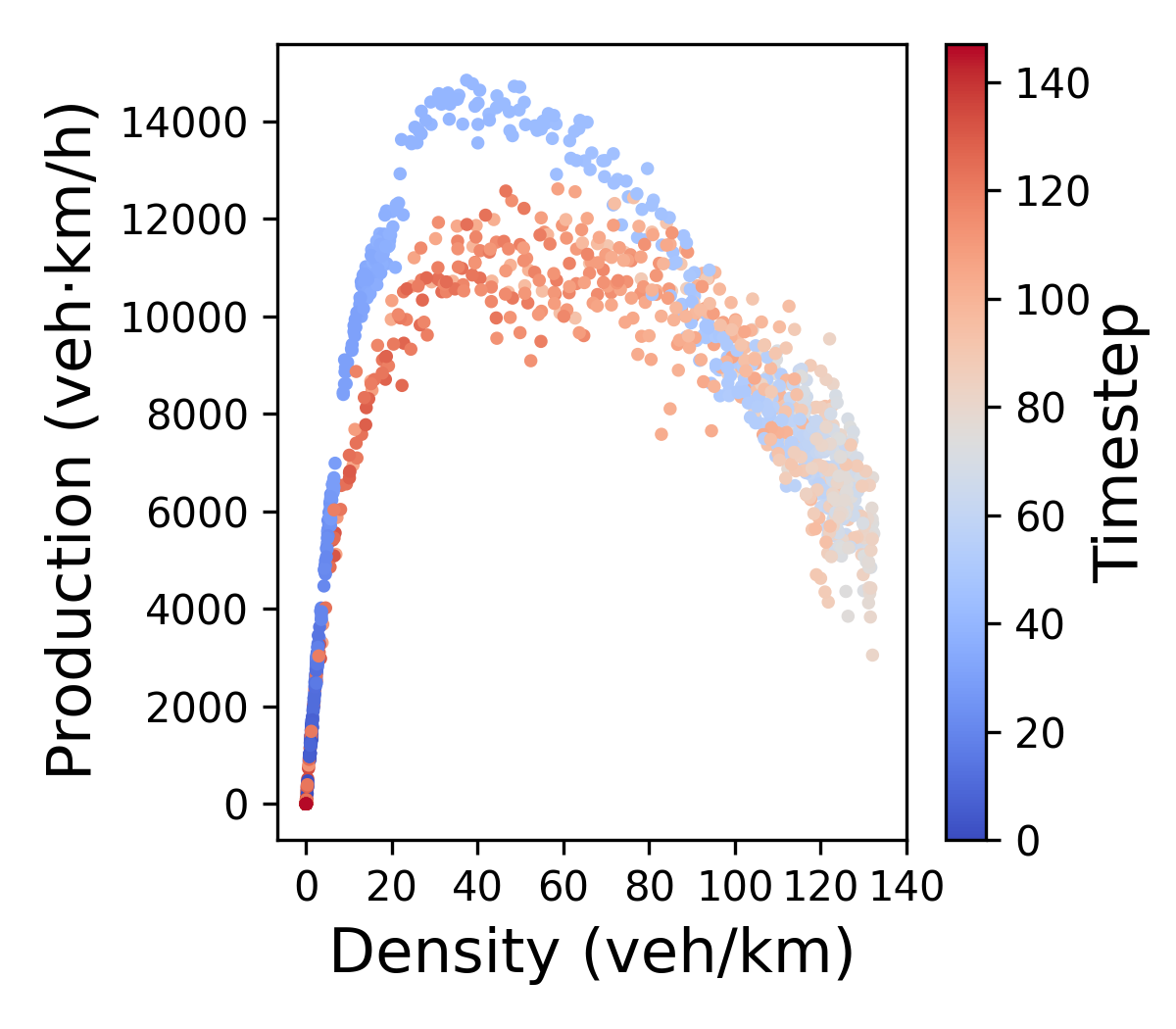}
    \caption{Loading and unloading of demand in NPC case.}
    \label{fig:mfd-loading}
  \end{subfigure}
  ~
  \begin{subfigure}[b]{0.48\columnwidth}
    \includegraphics[width=\linewidth]{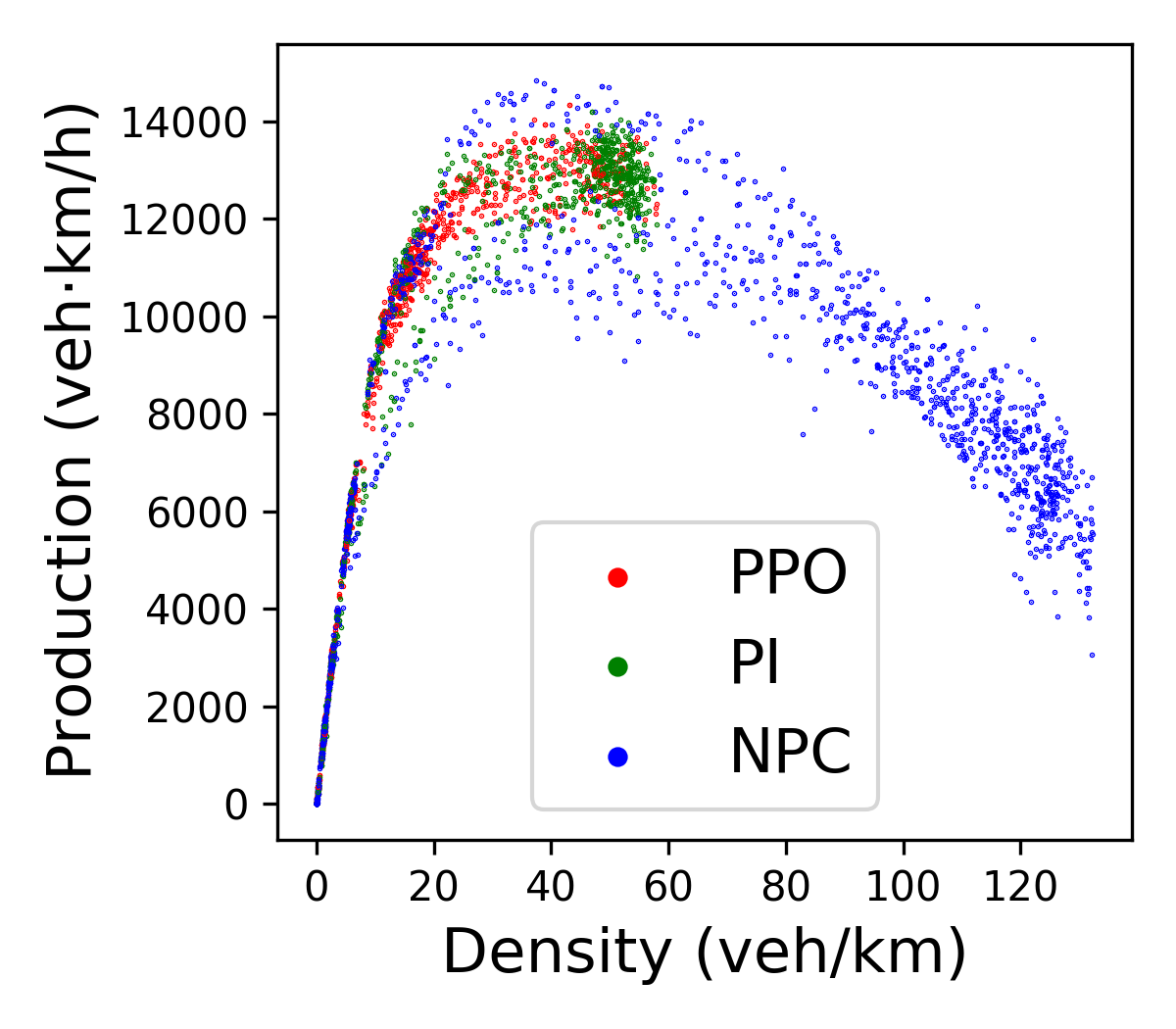}
    \caption{MFD with and without perimeter control.}
    \label{fig:mfd}
  \end{subfigure}
\caption{MFDs under different controllers.}
\end{figure}

\textbf{No Perimeter Control}:
Figure \ref{fig:mfd-loading} illustrates the MFD of the protected region under the processes of loading and unloading of demand, labeled with cool and warm colors, respectively. In the loading process, the traffic production increases monotonically when the protected region is unsaturated. However, when the density of the protected region exceeds the critical density of 35 $veh/km$, traffic production declines, reaching a minimum production of only one-third of the maximum production when the maximal attained density is four times the critical density. In the unloading process, hysteresis is present in the MFD indicating the capacity drop, which occurs when the traffic network recovers from extreme congestion.

\begin{figure}[htbp]
\centering
  \begin{subfigure}[b]{0.48\columnwidth}
    \includegraphics[width=\textwidth]{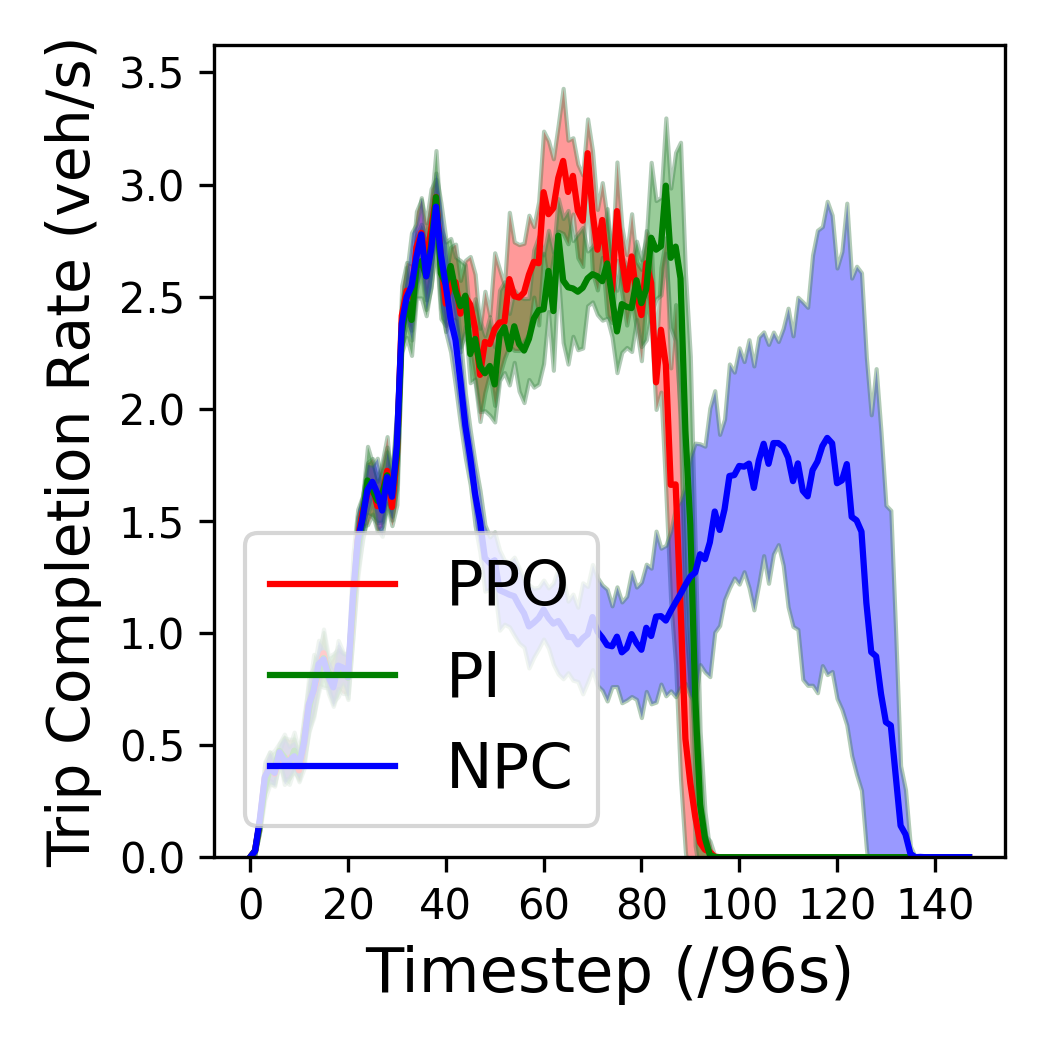}
    \caption{Trip completion rate.}
    \label{fig:tcr-vs-time}
  \end{subfigure}
    \hfill
    \begin{subfigure}[b]{0.48\columnwidth}
    \includegraphics[width=\textwidth]{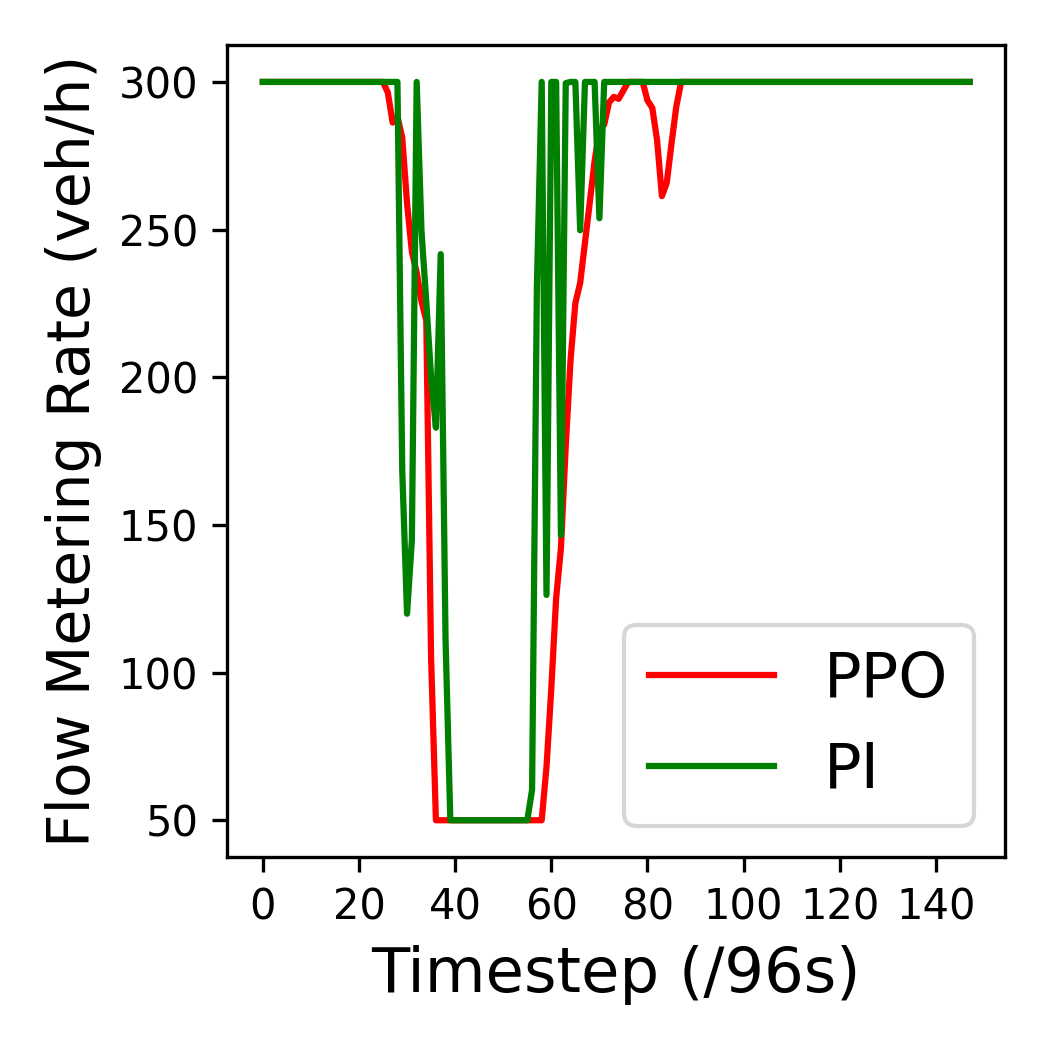}
    \caption{Flow metering rate.}
    \label{fig:action-vs-time}
  \end{subfigure}
\caption{Traffic metrics over time.}
\end{figure}

This efficiency loss is also evident in the trip completion rate over time shown in Figure \ref{fig:tcr-vs-time}. Before timestep 40, the rate gradually increases to around 3 $veh/s$, as the number of vehicles in the network rises. Between timestep 40 and 80, the rate decreases to approximately 1 $veh/s$. After timestep 80, the rate recovers from the loss; however, it does not recover to the previous rate, consistent with the production loss observed during the recovery process in the MFD. 

\vspace{-0.1cm}

\vspace{-0.1cm}

\textbf{PI Controller}:
The PI controller is reproduced from \cite{keyvan2012exploiting} based on our testbed with coefficients of proportional and integral terms being $K_P =36.88$ and $K_I=1.24$, and the set-point being $\hat{N}_2=2700 veh$. Variables $q_{in}(k)$, $N_2(k)$ in Eq. (\ref{eq:pi-control}) are the total flow metering rate at timestep $k$, number of vehicles in the protected region at timestep $k$, respectively. The flow metering rate is clipped with the same lower and upper bounds as the PPO controller. Note that the reproduced PI controller may not be optimal due to parameter design being vulnerable to uncertainties in the choice of set-point from the MFD and slope estimation. 
\begin{dmath}\label{eq:pi-control}
    q_{in}(k) = q_{in}(k-1) - K_{P}[N_2(k)-N_2(k-1)] + K_{I}[\hat{N}_2-N_2(k)]
\end{dmath}

\begin{figure}[htbp]
\centering
  \begin{subfigure}[b]{0.48\columnwidth}
    \includegraphics[width=\linewidth]{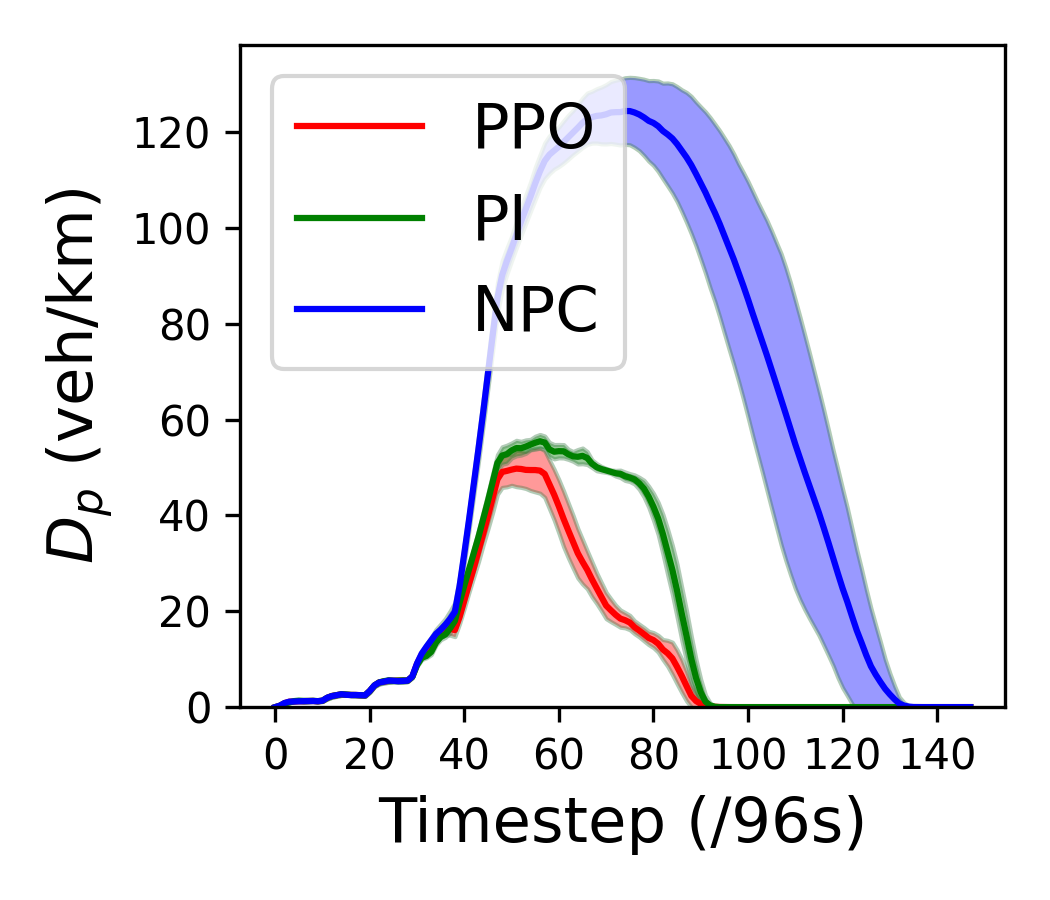}
    \caption{Inner density.}
    \label{fig:inner-density-vs-time}
  \end{subfigure}
  ~
  \begin{subfigure}[b]{0.48\columnwidth}
    \includegraphics[width=\linewidth]{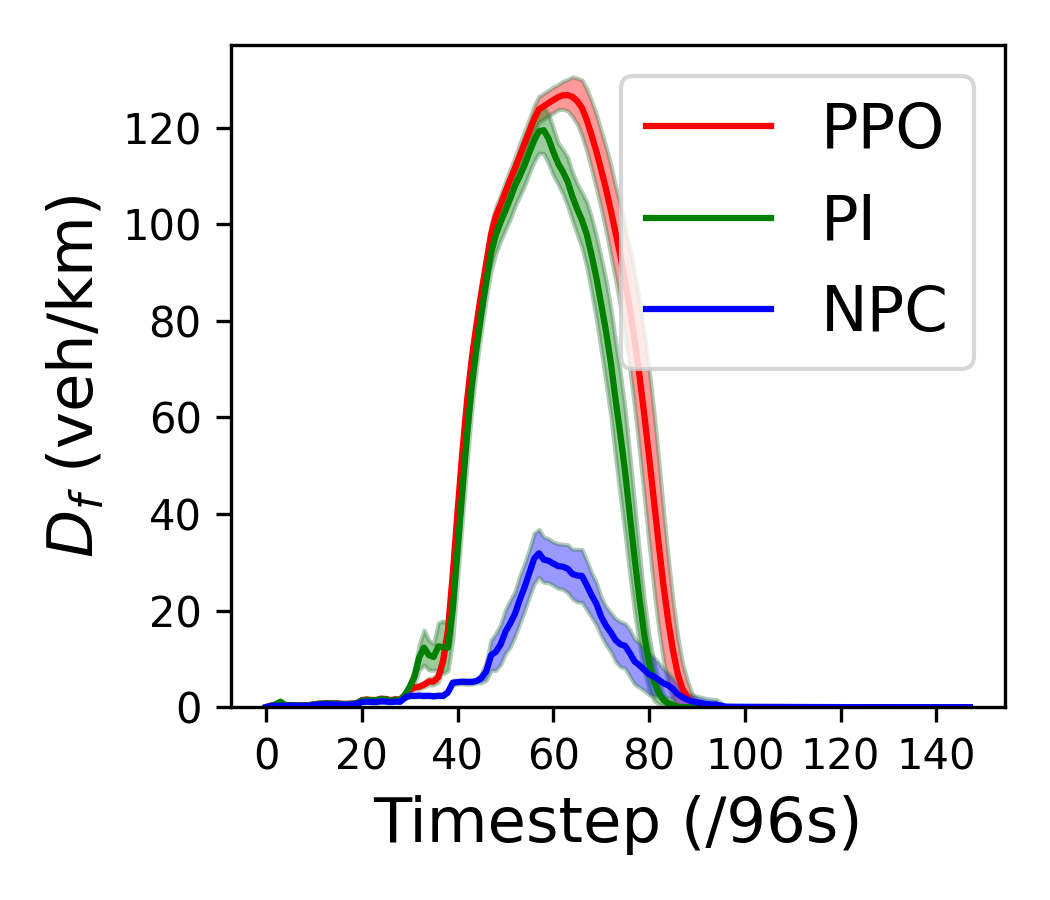}
    \caption{Feeder link density.}
    \label{fig:outer-density-incl-virtual-queue}
  \end{subfigure}
\caption{Congestion level over time.}
\end{figure}

Compared to NPC, the PI controller can keep the inner density near the critical density resulting in near-optimal traffic production, as is illustrated in Figure \ref{fig:mfd} of MFD and Figure \ref{fig:inner-density-vs-time} of inner density over time. For the trip completion rate, the PI controller maintains the same rate as NPC during the early simulation, while the PI controller has a higher rate than NPC between timestep 40 and 90. Note that the area under the curve is the total number of finished trips, the rate reaches zero earlier under the PI control than NPC, which implies the PI controller clears all trips earlier.

\textbf{PPO Controller}: In this part, the analysis of state space designs, and visualization of 1D, 2D and 6D state policies are provided. A comparison of controllers across various traffic metrics is also conducted to evaluate their effectiveness.

\begin{table*}
\centering
\footnotesize
\caption{Performance$^*$ of different state designs for the PPO controller.}
\begin{tabular}{>{\centering}p{2.5cm} |  >{\centering}p{0.8cm} >{\centering}p{1.0cm} >{\centering}p{0.8cm} >{\centering}p{0.8cm} >{\centering}p{1.0cm} >{\centering}p{0.8cm} >{\centering}p{0.8cm} >{\centering}p{0.8cm} >{\centering}p{0.8cm} >{\centering}p{0.8cm} >{\centering}p{0.8cm}} 
\hline
State\textbackslash Seed & 1 & 2 & 3 & 4 & 5 & 6 & 7 & 8 & 9 & 10 & Mean \tabularnewline
\hline
Inner Density Only & 5050 & 5640 & 5143 & 5181 & 4495 & 5356 & 5326 & 5968 & 5429 & 4998 & 5259 \tabularnewline\hline
Inner \& Feeder Link Densities & 4894 (3.1\%) & 5833 (-3.4\%) & 5064 (1.6\%) & \textbf{4793 (7.5\%)} & 4573 (-1.7\%) & 4934 (7.9\%) & 5273 (1.0\%) & 5801 (2.8\%) & 5134 (5.4\%) & 4802 (3.9\%) & 5110 (2.8\%) \tabularnewline\hline
Two Densities \& Future Demand & \textbf{4360 (13.7\%)} & \textbf{5151 (8.7\%)} & \textbf{5043 (1.9\%)} & 4901 (5.4\%) & \textbf{4218 (6.1\%)} & \textbf{4607 (14.0\%)} & \textbf{5069 (4.8\%)} & \textbf{5541 (7.2\%)} & \textbf{4801 (11.6\%)} & \textbf{4771 (4.5\%)} & \textbf{4846 (7.8\%)} \tabularnewline
\hline
\end{tabular}
\begin{tablenotes}[flushcenter]
\item[a] $^*$The performance is measured by the TTS (hour) and the percentage is the relative improvement to the 1D State case.
\end{tablenotes}
\label{tab:performance-comparison-state-design}
\end{table*}

\begin{table*}[!htbp]
\centering
\footnotesize
\caption{Performance$^*$ comparison of perimeter controllers on different replication seeds.}
\begin{tabular}{>{\centering}p{2.5cm} |  >{\centering}p{0.8cm} >{\centering}p{1.0cm} >{\centering}p{0.8cm} >{\centering}p{0.8cm} >{\centering}p{1.0cm} >{\centering}p{0.8cm} >{\centering}p{0.8cm} >{\centering}p{0.8cm} >{\centering}p{0.8cm} >{\centering}p{0.8cm} >{\centering}p{0.8cm}} 
\hline
Controller\textbackslash Seed & 1 & 2 & 3 & 4 & 5 & 6 & 7 & 8 & 9 & 10 & Mean \tabularnewline
\hline
NPC & 8614 & 14083 & 12471 & 12380 & 9537 & 13437 & 11800 & 14440 & 12359 & 11620 & 12074 \tabularnewline\hline
PI & 4840 (43.8\%) & 5706 (59.5\%) & 5458 (56.2\%) & 5328 (57.0\%) & 5332 (44.1\%) & 5410 (59.7\%) & 5664 (52.0\%) & 5680 (60.7\%) & 5459 (55.8\%) & 5299 (54.4\%) & 5417 (55.1\%) \tabularnewline\hline
PPO (Ours) & \textbf{4360 (49.4\%)} & \textbf{5151 (63.4\%)} & \textbf{5043 (59.6\%)} & \textbf{4901 (60.4\%)} & \textbf{4218 (55.8\%)} & \textbf{4607 (65.7\%)} & \textbf{5069 (57.0\%)} & \textbf{5541 (61.6\%)} & \textbf{4801 (61.1\%)} & \textbf{4771 (58.9\%)} & \textbf{4846 (59.9\%)} \tabularnewline
\hline
\end{tabular}
\begin{tablenotes}[flushcenter]
\item[a] $^*$The performance is measured by the TTS (hour) and the percentage is the relative improvement to the NPC case.
\end{tablenotes}
\label{tab:performance-comparison-npc-pi-ppo}
\end{table*}

\textbf{State Space Designs}:
Table \ref{tab:performance-comparison-state-design} presents the results of three state designs across 10 replications. Our states evolve in the order of increasing variable complexity and attainment difficulty from a traffic engineering perspective. By calculating the performance improvement relative to using $D_p$ only, including $D_f$ in the state enhances the performance in 8 out of 10 replications, with an average improvement of 2.8\%. Moreover, the incorporation of future demand into the state yields the best average performance among all state designs, with an average improvement of 7.8\%.

\begin{figure}[!htbp]
\centering
  \begin{subfigure}[htbp]{0.238\columnwidth}    \includegraphics[height=3cm, keepaspectratio]{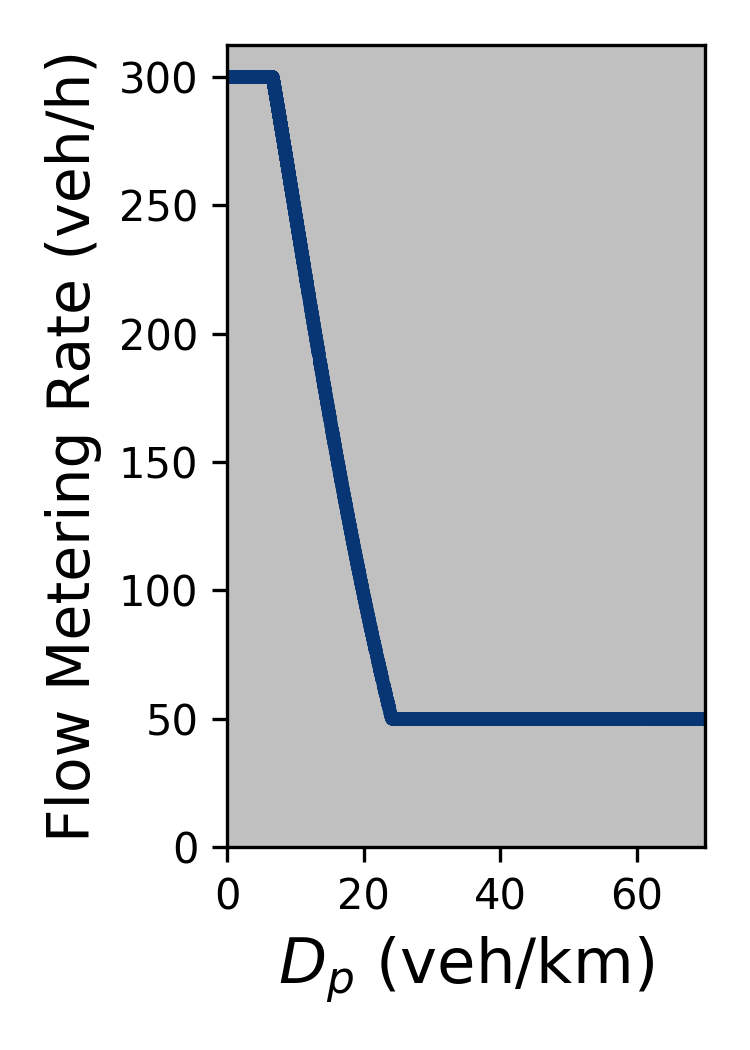}
    \caption{1D state.}
    \label{fig:policy-evolution}
  \end{subfigure}%
  \hfill
    \begin{subfigure}[htbp]{0.38\columnwidth}    \includegraphics[height=3cm, keepaspectratio]{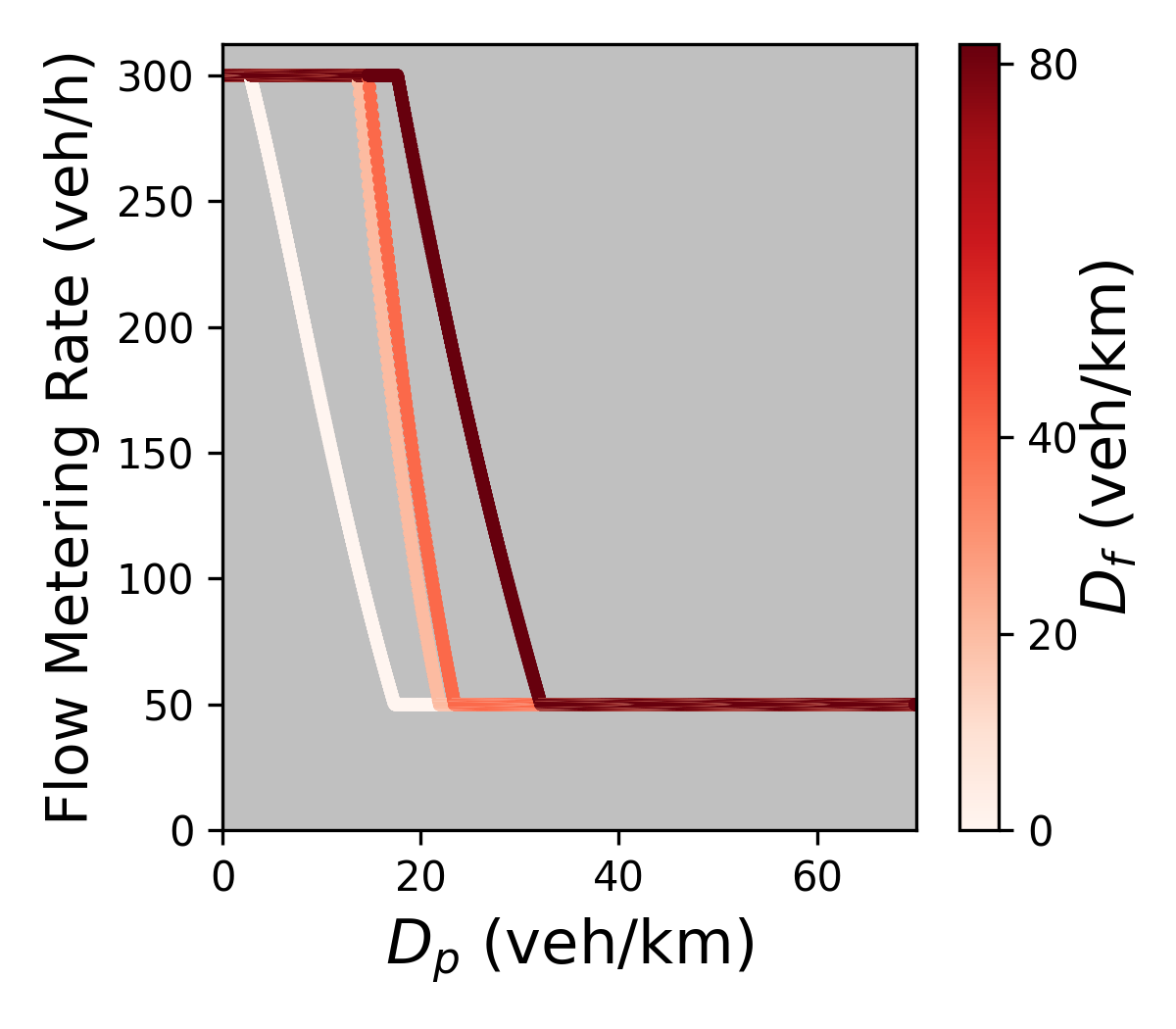}
    \caption{2D state.}
    \label{fig:2d-policy-visualization}
  \end{subfigure}%
  \hfill
    \begin{subfigure}[htbp]{0.38\columnwidth}
    \includegraphics[height=3cm, keepaspectratio]{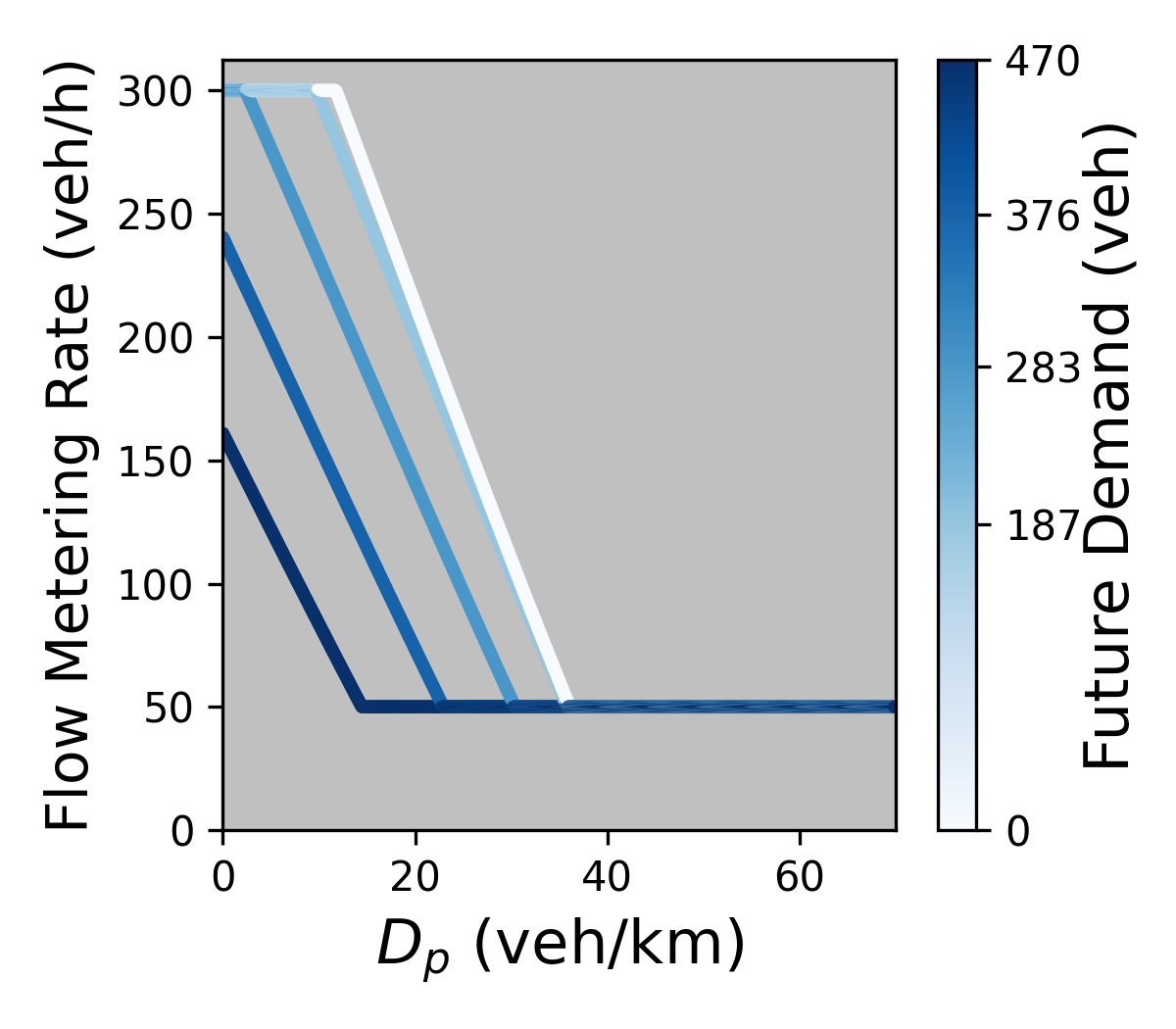}
    \caption{6D state.}
    \label{fig:6d-policy-visualization}
  \end{subfigure}
  \caption{Visualization of policies with various state designs.}
\end{figure}

\textbf{1D State Policy}: 
The simplest state design incorporates only the inner density $D_p$ as a one-dimensional state. The learnt 1D state policy is visualized in Figure \ref{fig:policy-evolution} exhibiting a three-piecewise linear function with 3 characteristics:

1) When $D_p$ is lower than 8 $veh/km$ being an undersaturated state, the policy employs the maximal metering rate to fully utilize the traffic network;

2) When $D_p$ exceeds 25 $veh/km$, signifying a near or oversaturated state, the policy applies the minimal metering rate to alleviate the internal congestion;

3) For $D_p$ between 8 $veh/km$ and 25 $veh/km$, the flow metering rate decreases smoothly and linearly as $D_p$ increases.

\textbf{2D State Policy}: The learnt policy is shown in Figure \ref{fig:2d-policy-visualization} with a color scheme representing $D_f$. The difference between 1D and 2D state policies is the 2D state policy enforces a higher flow metering rate with larger $D_f$ when the protected region is uncongested, allowing a less restricted perimeter control to fully utilize the traffic network.

\textbf{6D State Policy}: 
The learnt policy is visualized in Figure \ref{fig:6d-policy-visualization} with a color scheme representing the future demand. For the ease of visualization, the feeder link density $D_f$ is fixed to the average value across the simulation. The policy projection consists of several parallel piecewise linear functions resembling the 1D state policy, but shifted by future demand: For a fixed $D_p$, a larger future demand corresponds to a lower flow metering rate, logically resulting in stricter perimeter control when more vehicles are anticipated to be generated in both the protected region and feeder links. The most restrictive perimeter control, with a minimal flow metering rate of 50 $veh/h$, is implemented when the protected region is highly congested, even if the future demand is low.

In summary, the incorporation of feeder link density and future demand as state variables enhances the adaptiveness of our control policy as well as the performance in general.



\begin{figure}[!htbp]
    \centering
    \includegraphics[width=0.45\textwidth]{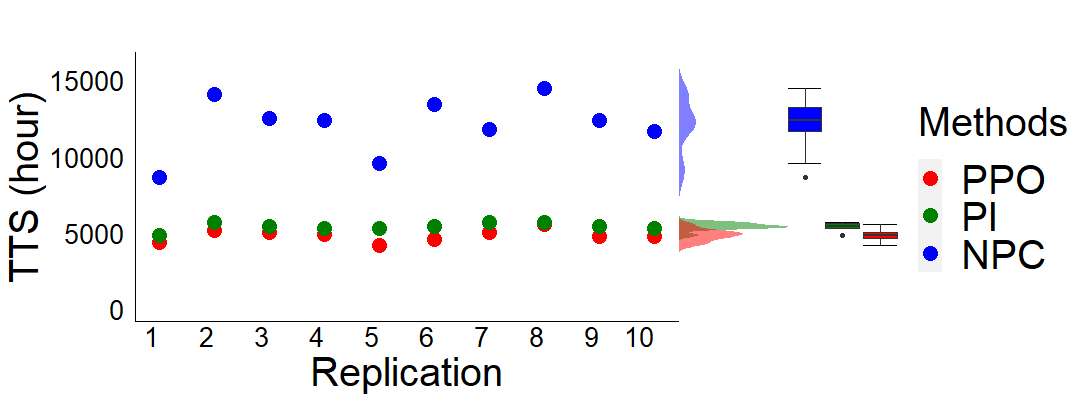}
    \caption{Performance comparison on every replication.}
    \label{fig:tts-vs-method-all-replication}
\end{figure}

\textbf{Comparison of Controllers}: The 6D state design for the PPO controller is chosen to compare with other controllers due to its outstanding performance among all state designs.
Figure \ref{fig:tts-vs-method-all-replication} displays the performance of all 10 replications: The scatter plot demonstrates that the PPO controller outperforms both the PI controller and the NPC in every replication. Furthermore, the kernel density estimation and box plot reveal a higher variability (low consistency) for the NPC case among all 10 replications. In contrast, the PI controller exhibits the lowest variability, although its average performance is slightly inferior to that of the PPO controller. The numerical results for Figure \ref{fig:tts-vs-method-all-replication} are presented in Table \ref{tab:performance-comparison-npc-pi-ppo}. On average, the PPO controller reduces the TTS by 59.9\% and 10.5\% compared to the NPC and PI controllers, respectively.



Figure \ref{fig:inner-density-vs-time} illustrates that the PPO controller maintains the lowest $D_p$ throughout the simulation, while NPC has the highest $D_p$. Figure \ref{fig:outer-density-incl-virtual-queue} indicates that the PPO controller exhibits the highest $D_f$ as the controller retains incoming vehicles outside the protected region to preserve a higher traffic production inside. As shown in Table \ref{tab:tts-inside-outside}, the reduced TTS inside the protected region is 9318 hours, which significantly surpasses the increased TTS on feeder links, totaling 2088 hours. Consequently, the PPO controller reduces the average TTS by 7220 hours compared to the NPC case.
\vspace{-0.4cm}
\vspace{-0.4cm}

\begin{table}[htbp]
\centering
\caption{TTS (hour) inside \& outside protected region.}
\label{tab:tts-inside-outside}
\begin{tabular}{c|c|c|c|}
\cline{2-4}
                                 & TTS inside & TTS outside & TTS \\ \hline
\multicolumn{1}{|c|}{PPO}        &  2425         & 2422           & 4847   \\ \hline
\multicolumn{1}{|c|}{NPC}        & 11743          & 334           & 12077   \\ \hline
\multicolumn{1}{|c|}{Difference} & -9318          & 2088          & -7220   \\ \hline
\end{tabular}
\end{table}





In Figure \ref{fig:action-vs-time}, both the PPO and PI controllers exhibit highly dynamic actions over time for replication 1. The PI controller's action oscillates considerably, while the PPO controller displays a much smoother action progression. Additionally, between timestep 20 and timestep 90, the PPO controller employs lower actions than the PI controller to have a more restricted control.

Figure \ref{fig:tcr-vs-time} presents the trip completion rate under different perimeter controllers. Before timestep 38, the rates of the PPO and PI controllers are nearly identical to the NPC case since the entire network is loading demand from an empty state. Between  timestep 38 and timestep 85, the PPO controller achieves a higher rate than the PI controller and reaches a zero rate earliest, indicating that all trips are completed. However, The NPC has a 50\% lower rate compared to either PPO or PI controllers and all trips are only completed near timestep 140. Additionally, the variability for the NPC case is the highest among the compared controllers.





\begin{figure}[!htbp]
\centering
  \begin{subfigure}[b]{0.48\columnwidth}
    \includegraphics[width=\linewidth]{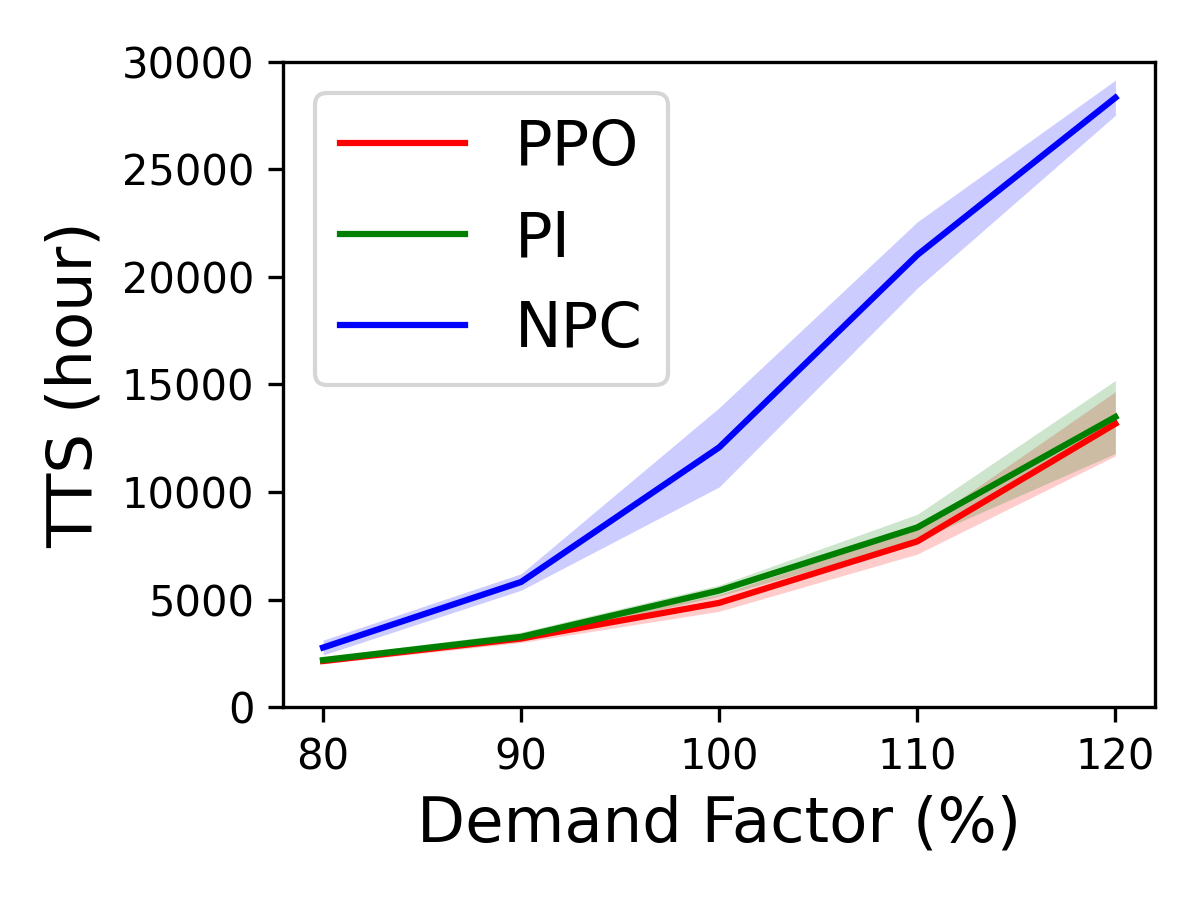}
    \caption{Scalability test.}
    \label{fig:scalability}
  \end{subfigure}
  ~
  \begin{subfigure}[b]{0.48\columnwidth}
    \includegraphics[width=\linewidth]{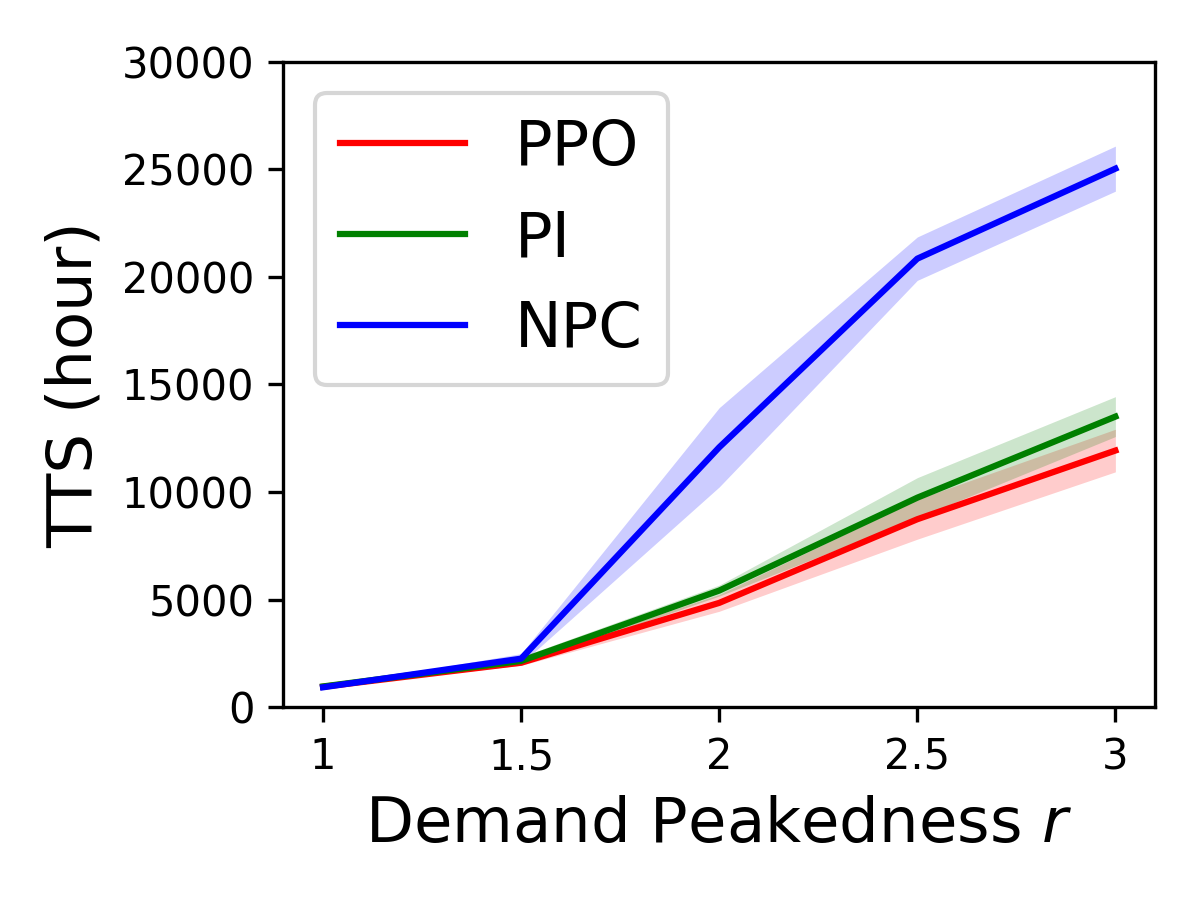}
    \caption{Peakedness test.}
    \label{fig:generalizability}
  \end{subfigure}
\caption{Generalizability tests for robustness.}
\end{figure}

\textbf{Generalizability Tests}: The objective of the generalizability tests is to evaluate the controller's performance under out-of-distribution scenarios not experienced during training. 
Two generalizability tests are performed: 1) a scalability test scales the total number of vehicles up or down while maintaining a fixed peakedness; 2) a demand peakedness test varies the demand shape from a flat one to a highly peaked one while keeping the total number of vehicles constant. Results are presented in figures \ref{fig:scalability} and \ref{fig:generalizability} correspondingly. 
\vspace{-0.15cm}
\vspace{-0.6cm}

In the scalability test, the demand is adjusted with factors ranging from 80\% to 120\%, incremented by 10\%. Each controller is trained exclusively on the 100\% factor. In Figure \ref{fig:scalability}, the PPO controller achieves the lowest average TTS, followed by the PI controller and NPC. Moreover, the performance gap between the PPO controller and NPC widens as the demand factor increases, indicating that the benefits of perimeter control are more significant when faced with a more congested scenario. This suggests that perimeter control is beneficial only when a  traffic network experiences production degradation due to congestion; otherwise, implementing perimeter control is unnecessary.

In the peakedness test, the demand profile's peakedness is changed with the parameter $r$ from 1.0 to 3.0 in increments of 0.5. All controllers are trained exclusively on demand peakedness $r=2$. In Figure \ref{fig:generalizability}, when the demand profile is entirely or nearly uniform over time ($r=1.0$ and $r=1.5$), the performances of the three controllers are quite similar, indicating that perimeter control may not be needed. As $r$ increases beyond 1.5, the performance gap between the NPC and both the PPO and PI controllers widens with the peakedness. Moreover, the PPO controller consistently outperforms the PI controller in all tested scenarios, demonstrating that the PPO controller remains robust under changes in demand peakedness relative to the baselines. 


The performance disparity between the PI  controller and the PPO controller is rooted in their characteristics. The PI controller is a regulator, while the PPO controller is an optimal controller. The PI controller does not explicitly strive for optimality and relies on an accurately identified set-point. If this set-point is not precise, the performance of the PI controller may not match that of the PPO controller.


\section{Conclusion and Future Work}

We have explored an RL-based homogeneous flow rate optimization for perimeter control that does not require NTMs or MFDs. The evaluation of control policies is performed with a grid-patterned traffic network.
In addition, the choice of each state variable is supported by an in-depth policy visualization analysis. Moreover, our model-free approach matches a standard model-based PI controller and proves to be more robust under scalability and demand peakedness tests.  
The exploration of the characteristics and requirements of a model-free approach demonstrates that it is more adaptive to uncertainty and hence less susceptible to model inaccuracies.

Future research directions include model-free heterogeneous flow rate optimization for enhanced perimeter queue balancing and management, the replacement of flow meterings with  traffic signals on the perimeter, and an in-depth comparison of model-based RL perimeter control methods against model-free ones.

\bibliographystyle{IEEEtran}
\bibliography{refs}

\end{document}